\documentclass[11pt,a4paper]{article}
\usepackage[hyperref]{acl2021}

\usepackage{microtype}
\usepackage{graphicx}
\usepackage{booktabs} 
\usepackage{graphicx}
\usepackage{url}
\usepackage{amsmath}
\usepackage{makecell}
\usepackage{float}
\usepackage{enumitem}
\usepackage{url}
\usepackage{soul}
\usepackage{multirow}
\usepackage{mathtools,kantlipsum}
\usepackage{amssymb}
\usepackage{color, colortbl}
\usepackage{inconsolata}
\usepackage{enumitem} 
\usepackage[size=scriptsize]{todonotes}

\usepackage{times}
\usepackage{latexsym}

\usepackage{microtype}

\aclfinalcopy 
\def\aclpaperid{182}

\newcommand{\my}[1]{{\fontfamily{pcr}\selectfont #1}}
\newcommand{\ours}[0]{ATT}

\title{An Adversarially-Learned Turing Test\\for Dialog Generation Models}

\author{Xiang Gao \quad\quad Yizhe Zhang  \quad\quad \textbf{Michel Galley} \quad\quad \textbf{Bill Dolan}\\
  Microsoft Research, Redmond, WA, USA \\
  {\small \tt \{xiag,yizzhang,mgalley,billdol\}@microsoft.com}
}
\date{}

\begin{document}
\maketitle

\begin{abstract}
The design of better automated dialogue evaluation metrics offers the potential of accelerate evaluation research on conversational AI. However, existing trainable dialogue evaluation models are generally restricted to classifiers trained in a purely supervised manner, which suffer a significant risk from adversarial attacking (\textit{e.g.}, a nonsensical response that enjoys a high classification score).
To alleviate this risk,
we propose an adversarial training approach to learn a robust model, \ours{} (Adversarial Turing Test), that discriminates machine-generated responses from human-written replies. In contrast to previous perturbation-based methods, our discriminator is trained by iteratively generating unrestricted and diverse adversarial examples using reinforcement learning. The key benefit of this {\it unrestricted} adversarial training approach is allowing the discriminator to improve robustness in an iterative attack-defense game.
Our discriminator shows high accuracy on strong attackers including DialoGPT and GPT-3.\footnote{Code and model be open-sourced on \url{https://github.com/golsun/AdversarialTuringTest}.}

\end{abstract}

\section{Introduction}
\label{sec:intro}

Turing Test \cite{turing1950computing} was proposed to assess whether a machine can think. A machine and a human player communicate with a human judge and try to convince the judge that they are the human.
This test provides an evaluation framework -- a machine is intelligent to certain extent if it passes the Turing Test.
To allow fast and less expensive evaluations, the human judge is often replaced by an automated human-vs-machine classifier \cite{von2003captcha, baird2003pessimalprint, rui2004artifacial, lowe2017towards}.

An automated Turing test is straightforward for constrained scenarios with unambiguous correct answer, such as text classification. 
In contrast, for open-domain conversation, an infinite number of plausible responses for the same context exist, and they may differ from each other substantially. 
In this case, the existing automated evaluation methods \cite{papineni2002bleu, zhang2019bertscore, sellam2020bleurt} which measure hypothesis quality by its similarity with reference answers become sub-optimal, because it is difficult to find a set of diverse reference answers to cover such one-to-many possibilities in dialogue.
Furthermore, it is impossible to use these reference-based metrics in scenarios when the reference is not available, e.g., online chatbots.

These challenges motivate an alternative approach, \textit{i.e.}, trainable reference-free metrics \cite{albrecht2007re, guan2020union, gao2020dialogrpt}.
Previous works generally frame the task as a supervised learning (SL) problem, training a classifier to distinguish human and machine outputs, or a regression model to fit the human ratings.
However, trainable metrics have potential problems of being gamed using adversarial attacking \cite{albrecht2007re, sai2019re, gao2019neural}. 

To learn a more robust evaluation metric, we propose to train a model to discriminate machine outputs from human outputs via iterative adversarial training, instead of training evaluation model with a fixed dataset.
In contrast to previous perturbation-robust methods that only modify characters or words \cite{ebrahimi2017hotflip, li2018textbugger, gao2018black}, we generate ``unrestricted'' adversarial examples by fine-tuning a dialogue response generator to maximize the current discriminator score via reinforcement learning. 
This is followed by training the discriminator that accounts for these additional adversarially generated examples. The above two steps are repeated as an iterative attack-defense game until the generator no longer can decrease the discriminator accuracy below a threshold.

To further improve the robustness of the discriminator, we reduce the chance that the discriminator be fooled by unseen patterns by increasing the diversity of the adversarial examples in several novel ways. 
Firstly, we explicitly encourage the adversarial dialogue responses to be context-sensitive by including rewards from a pre-trained context-response matching model. 
Secondly, we decode with different decoding settings when generating adversarial examples. 
Our discriminator showed high accuracy on several strong attackers including DialoGPT \cite{zhang2019dialogpt} and GPT-3 \cite{brown2020gpt3}.

\section{Method}
We define the problem as learning a discriminator to distinguish machine-generated and human-written responses for open-domain dialogue context.
Similar to training a generative adversarial network (GAN) \cite{goodfellow2014generative}, our \my{\ours{}} (\my{A}dversarial \my{T}uring \my{T}est) method involves two sets of competing components: discriminators \my{D} that defend, and generators \my{G} that attack.

\subsection{Discriminator}
A supervised learning (SL) approach is employed to train the discriminator. Following \citet{gao2020dialogrpt}, the loss is defined to increase the probability of picking the human-written responses $y^H$ when mixed with machine-generated hypotheses $y^M$ for the same context $x$.
    \begin{align}  
        \mathcal{L}_{SL}(\theta_D) = - \sum_{i} \log{\frac{
            e^{h(x_i, y^H_i)}
            }{
            e^{h(x_i, y^H_i)} + e^{h(x_i, y^M_i)}
            }}
    \end{align}
where $h(x,y,\theta_D)$ is the scalar output from the discriminator, which is parameterized by $\theta_D$. 
At inference time, we compute the score 
    \begin{align}
        s(y|x) = \text{Sigmoid}(h(x,y))
    \end{align}
The discriminator is implemented by adding a linear layer to GPT-2 transformers \cite{radford2019gpt2}, following \citet{gao2020dialogrpt}. 

\subsection{Generator}

We generate 
adversarial examples via a generator \my{G} trained with reinforcement learning (RL) using policy gradient \cite{williams1992reinforce}. 
For each context $x$, the generator generates $n$ hypotheses $\{y_i\}$. The reward $R(y_i)$ is defined with a baseline $b(x)$, which is used to reduce the variance of gradients:
    \begin{align}
        R(y_i) = s(y_i|x) - b(x) \\
        b(x) = \frac{1}{n} \sum_{j=1}^{n} s(y_j|x) 
    \end{align}
Applying Policy Gradient \cite{williams1992reinforce}, we minimize the following loss: 
    \begin{align}
        \mathcal{L}_{RL}(\theta_G) = - \sum_{i=1}^{n} \log P(y_i|x,\theta_G) R(y_i) 
    \end{align}
where $P(y|x,\theta_G)$ is the probability generating $y$ given $x$ from the generator parameterized by $\theta_G$.
The generator is implemented using a GPT-2 architecture \cite{radford2019gpt2}, following \citet{zhang2019dialogpt}.

\subsection{An iterative attack-defense game}

\begin{figure}[t]
    \centering
    \includegraphics[width=0.47\textwidth]{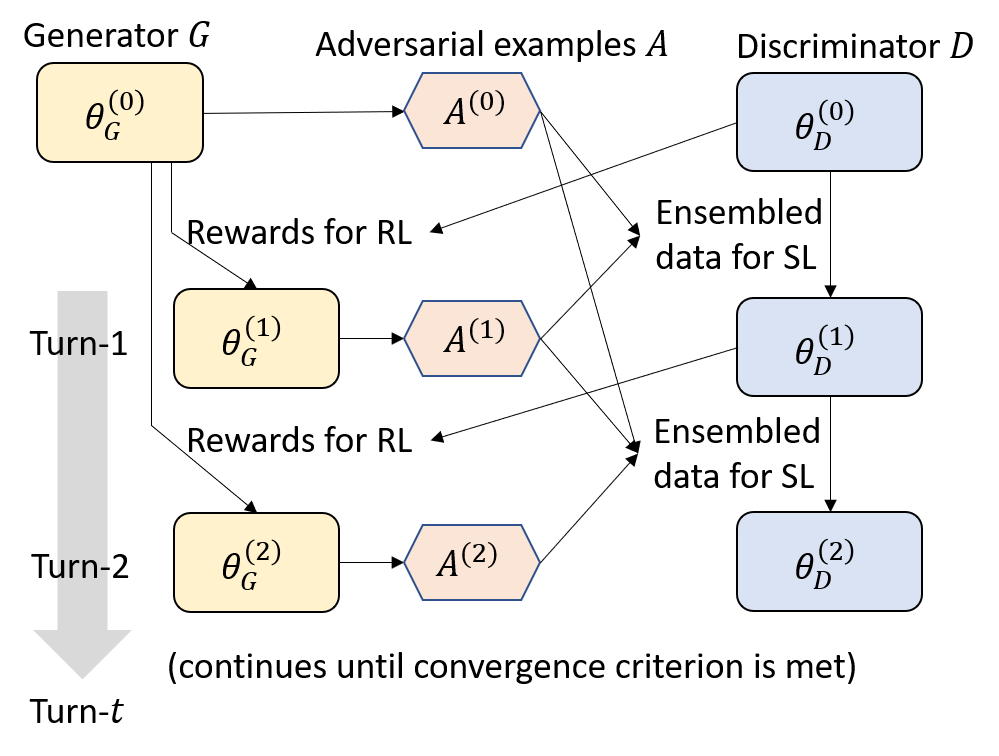}
    \caption{Illustration of the first several turns of the iterative attack-defense game between the discriminator, trained with supervised learning (SL), and generator, trained with reinforcement learning (RL). }
    \label{fig:iterative_game}
\end{figure}

We first pre-train the components individually and then jointly train them in an iterative attack-defense game, as illustrated in Figure~\ref{fig:iterative_game}.

\my{HvM} is initialized using, $\theta_D^{(0)}$, the weights of a human-vs-machine classifier from \citet{gao2020dialogrpt} trained in a SL manner to classify whether a response is a human response or DialoGPT generated.

Each turn of the game starts with an attack phase that
trains \my{G} with RL to attack \my{D}.
The training is stopped when the validation accuracy of \my{D} is dropped below threshold $c_\text{low}$, or the training steps exceed certain number $N_G$, whichever comes first.
The turn then switches to a defense phase. \my{D} is trained using samples generated from \my{G} via SL. We stop training the discriminator when the validation accuracy is higher than threshold $c_\text{hi}$, or the training steps exceed certain number $N_D$. 
We repeat this process until the validation accuracy of \my{D} in the last $m$ turns is always kept higher 
than $c_\text{hi}$. 

Note above procedure resembles GAN \cite{goodfellow2014generative}. However, we focus on improving the robustness of the classifier, while GAN focus on generating realistic examples. From a modeling perspective, the key differences with a conventional GAN are: 
\label{sec:diff}

\paragraph{Generator re-initialization} We set \my{G} as $\theta_G^{(0)}$ at the beginning of each attack phase, instead of continuing to learn from the last turn. This is because, at the early stage of the game, \my{G} often learns to generate adversarial examples that are not fluent or grammatically correct (one example at Turn-5 is shown in Table~\ref{table:eg}). They can successfully fool \my{D}, as initially \my{D} has not seen such adversarial examples. 
However, such attacker \my{G}, which typically learns  disfluent adversarial examples in earlier iterations, is difficult to fine-tune towards well-formed adversarial examples. 
To allow the late-stage attackers to generate fluent adversarial examples, we reset \my{G} parameters as the pre-trained generator before each attack phase to obtain multiple attackers as shown in Figure~\ref{fig:iterative_game}. We empirically find that such a strategy gives better performance than initializing each attacker with the previous one as in GAN.

\paragraph{Adversarial example ensemble} We train \my{D} on samples generated by \my{G} from all previous turns, instead of only the last turn. As shown in Figure~\ref{fig:iterative_game}, at Turn-2, \my{D} is trained on samples from pre-trained \my{G}, $A^{(0)}$, samples of of Turn-1 $A^{(1)}$, and Turn-2 $A^{(2)}$. At each turn, we stop training \my{D} when the validation accuracy of all these datasets $A^{(t)}$ is higher than threshold $c_\text{hi}$. This enables the defender to capture all observed attacks rather than only focusing on the adversarial examples generated by the last attacker as in GAN, thus yielding a more robust defense against various attacks.

\subsection{Diversifying the adversarial examples}
\label{sec:div}

We find that mode collapse happens when only using a single human-vs-machine discriminator \my{HvM}. That is, the generated adversarial examples tend to be insensitive to the context, indicating that the generator finds a universal adversarial attacking pattern that can successfully attack \my{HvM} for most contexts. However, a corpus of similar adversarial examples makes adversarial training inefficient and the discriminator less robust. Therefore, we encourage the content of the adversarial examples to be diverse by integrating \my{HvM} and a pre-trained human-vs-random classifier \cite{gao2020dialogrpt} \my{HvR}. \my{HvR} is trained to predict whether a response is randomly retrieved or is the ground truth. The final discriminator score $s_D(y|x)$ is the geometric mean\footnote{Geometric mean has an advantage over the arithmetic mean in that it encourages both $s_\text{HvM}$ and $s_\text{HvR}$ to be high. The final value is zero if one of them is zero.} of the outputs of \my{HvM} and \my{HvR}:
    \begin{align}
        s(y|x) = \sqrt{s_\text{HvM}(y|x) \, s_\text{HvR}(y|x) }
        \label{eq:ensemble}
    \end{align}
    
For the generator, we increase the diversity of adversarial examples by randomly changing the hyper-parameter of the generator decoding process. The decoding temperature $T$ is uniformly sampled from a range of levels (0.3, 1, 10, 100), to control the token generation probability distribution.

\section{Experiments}

\subsection{Data}
As we focus on open-domain dialogue, we use Reddit data obtained from a third-party Reddit dump,\footnote{https://files.pushshift.io/reddit/} following \citet{gao2020dialogrpt}.  

\subsection{Baselines}
\label{sec:baseline}
We compare \my{\ours{}} with the following models:
\begin{itemize}[nosep]
    \item \my{SL} by \citet{gao2020dialogrpt} is trained via SL on human-vs-DialoGPT data. 
    \item \my{GAN} is similar to \my{\ours{}}, 
    but
    it does not apply the generator re-initialization or adversarial example ensemble strategies 
    of Section~\ref{sec:diff}.
    \item \my{ND} (Non-Diverse \ours{}) a variant to \my{\ours{}}, without the diversifying objective in Section~\ref{sec:div} to diversify adversarial examples. 
\end{itemize}

\subsection{Results}

\begin{table*}[ht]
    \centering
    \small
    \begin{tabular}{
    p{0.2\textwidth} p{0.7\textwidth}
    }
    \Xhline{2\arrayrulewidth}
    \textbf{Context} & Why does anything exist? \\
    \hline
    
    Human & Well, some philosophers have dealt with it, but I have never left satisfied.  \\
    Parrot             &  Why does anything exist?\\
    DialoGPT, greedy           & I'm not sure what you're trying to say.\\
    DialoGPT, sampling           & Seems like a good debate topic.\\
    GPT-3, greedy      & I don't know. \\
    GPT-3, sampling    & A question like this is beyond my capability to solve. \\
    Adversarial \my{G}, turn-5    & Thou doth not know Jesus his brother, and thou shalt thank him over zeth thou thirstin. \\
    Adversarial \my{G}, turn-40    & Man, as an Buddhist, that answer is, in fact, no. Inside out there is nothing. Edits up. \\
    
    \Xhline{2\arrayrulewidth}
    \end{tabular}
    \caption{Examples of responses generated from different attackers for the same context.}
    \label{table:eg}
\end{table*}

\begin{figure}[t]
    \centering
    \includegraphics[width=0.5\textwidth]{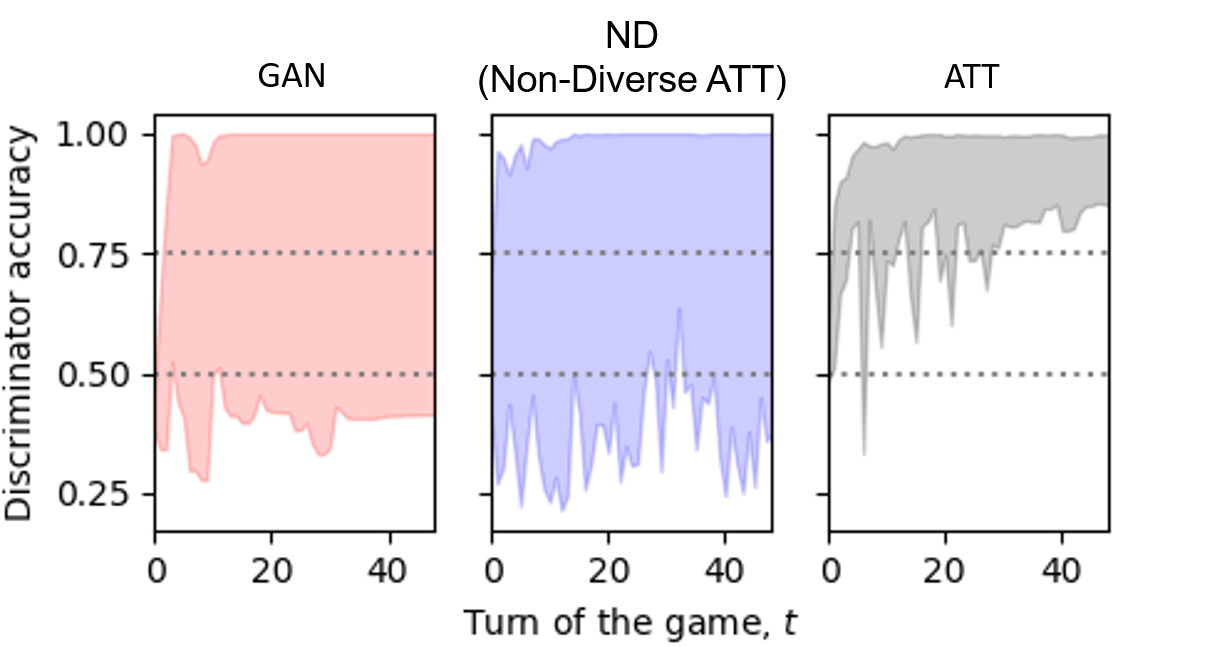}
    \caption{Discriminator accuracy after attack phase at each turn of the game. Filled area shows its range over past adversarial datasets $\{A^{(i)}\}$, $i=0 \sim t$. 
    }
    \label{fig:converged}
\end{figure}

As shown in Figure~\ref{fig:converged}, the accuracy of \my{D} for \my{\ours{}} method gradually increases as the game continues, and finally remains above 0.75 after about 30 turns of the game. This indicates our convergence criterion is met.
The convergence of \my{D} accuracy is accompanied by the improvement of \my{G} generation quality.
As shown by the examples of Table~\ref{table:eg}, a response that is not grammatically correct in the early stage (e.g., turn-5) can successfully fool \my{D}, but \my{G} tends to generate more human-like response
at later stages (e.g., turn-40). This is desirable as an ideal discriminator should only fail when generated responses are sufficiently similar to human-written replies. 

For \my{GAN}, its discriminator accuracy has a wide variance band 
and always performs poorly for some attackers (accuracy $<$ 0.5 in Figure~\ref{fig:converged}).
It is because, although its discriminator successfully defends against its latest generator, it forgets the pattern learned from previous turns. 

For \my{ND}, 
the adversarial examples tend to be similar for different contexts, which makes learning inefficient. Therefore, its discriminator can be easily attacked with a new generator. This makes its accuracy lower than 0.50, as shown in Figure~\ref{fig:converged}.

Besides DialoGPT and its adversarially-trained version, Adversarial \my{G}, we consider the following external attackers to test the robustness of the learned discriminator. See Table~\ref{table:eg} for examples. 
\begin{itemize}[nosep]
    \item \my{GPT-3} \cite{brown2020gpt3}: We called its online API to obtain its dialogue responses, in both greedy and sampling decoding settings. 
    \item \my{Parrot}: This system samples a turn from the context as the response, which is very relevant to the context (high \my{HvR} score) but are generally not good ones. We consider this system to test whether \my{D} solely rely on relevancy.
\end{itemize}
As illustrated in Table~\ref{table:acc}
, 
\my{SL} 
shows poor performance on many unseen datasets.
\my{GAN} and \my{ND} perform slightly better on \mbox{GPT-3} but not well on other attacks. In contrast, \my{\ours{}} shows a significant increase in accuracy even for unseen \mbox{GPT-3} datasets.

\begin{table}[ht]
    \centering
    \small
    \begin{tabular}{
    p{0.18\textwidth} | p{0.04\textwidth} p{0.04\textwidth} p{0.04\textwidth} p{0.04\textwidth}
    }
    \Xhline{2\arrayrulewidth}
     \multirow{2}{*}{Attacker} & \multicolumn{4}{c}{Defender} \\
     & SL & GAN & ND & \ours{} \\
    \hline
    
    Parrot                    & \cellcolor{blue!24}\textbf{0.977} & \cellcolor{blue!16}0.816 & \cellcolor{blue!3 }0.550 & \cellcolor{blue!23}0.967 \\
    DialoGPT, greedy          & \cellcolor{blue!15}0.807 & \cellcolor{blue!2 }0.544 & \cellcolor{blue!7 }0.632 & \cellcolor{blue!19}\textbf{0.877} \\
    DialoGPT, sampling        & \cellcolor{blue!0 }0.458 & \cellcolor{blue!0 }0.482 & \cellcolor{blue!5 }0.597 & \cellcolor{blue!10}\textbf{0.695} \\
    Adversarial \my{G}, worst & \cellcolor{blue!1 }0.518 & \cellcolor{blue!0 }0.410 & \cellcolor{blue!0 }0.350 & \cellcolor{blue!17}\textbf{0.853} \\
    GPT-3, greedy             & \cellcolor{blue!10}0.699 & \cellcolor{blue!14}0.780 & \cellcolor{blue!11}0.732 & \cellcolor{blue!17}\textbf{0.857} \\
    GPT-3, sampling           & \cellcolor{blue!0 }0.460 & \cellcolor{blue!5 }0.589 & \cellcolor{blue!2 }0.546 & \cellcolor{blue!7 }\textbf{0.626} \\
 
    \Xhline{2\arrayrulewidth}
    \end{tabular}
    \caption{Accuracy of the discriminators (defenders). Darker cell color indicates better performance.}
    \label{table:acc}
\end{table}

\label{sec:related}
\section{Related Work}

\paragraph{Dialogue evaluation and ranking.}
Open-domain dialogue systems are often evaluated using similarity between hypotheses and reference, e.g. BLEU \cite{papineni2002bleu}. \citet{lowe2017adem} trained an evaluation model with context, reference, and hypothesis as inputs. Complementary to this, corpus-level metrics for diversity \cite{li2016mmi, zhang2018gan} and other aspects are proposed. When reference is not available, dialogue ranking models \cite{xiaoice, gao2020dialogrpt} are often employed, mostly trained via SL. \citet{dinan2019build} trained a toxic dialogue classifier with human in the loop to provide adversarial examples.

\paragraph{Reinforcement Learning.} RL has been used to guide the dialogue generator using rewards, which can be hand-designed \cite{li2016rl}, obtained from a pre-trained classifier \cite{shin2019happybot}, or extracted from user response \cite{jaques2020offline}.

\paragraph{Adversarial attack and defense.}
Most existing works create adversarial examples by adding perturbation in embedding space \cite{miyato2016adversarial, zhao2017generating}, by editing characters \cite{ebrahimi2017hotflip}, or tokens \cite{alzantot2018adversarial, gao2018black}.
Adversarial training \cite{biggio2013evasion, szegedy2013intriguing, goodfellow2014explaining} is then used to improves the model robustness. 
Unrestricted adversarial examples is a relatively less studied field, and most works are for images \cite{brown2018unrestricted, song2018unrestricted, wang2019nonconstrained, qiu2020semanticadv}.

\section{Conclusions}

In this work we propose to learn a robust human-vs-machine discriminator from an iterative attack-defense game. Diversified and unrestricted adversarial examples are automatically generated and used to fine-tune the discriminator. It significantly increased accuracy and robustness in terms of classification accuracy on unseen attacks.

\section*{Ethical Considerations}

We cautiously advise users of our system be careful about the potential bias in the dataset used to train our model.
The raw data is publicly available, but the texts written by human have varying levels of quality. The dataset may contain offensive and/or toxic language. A proper definition of human-written text quality is beyond the scope of this work, as we focus on learning a human-vs.-machine discriminator in this short paper.

We used one P100 GPU for training. The training time for each method is approximately 48 hours. The generator and discriminator have about 700M parameters. The code and our models will be open-sourced together with the details of training hyperparameters.

\clearpage

\bibliography{acl2021}
\bibliographystyle{acl_natbib}

\end{document}